\documentclass[letterpaper, 10 pt, conference]{ieeeconf}

\IEEEoverridecommandlockouts                              % This command is only needed if 
\usepackage{times}
\usepackage{epsfig}
\usepackage{graphicx}
\usepackage{amsmath}
\usepackage{amssymb}
\usepackage{subfigure}
\usepackage{csquotes}
\usepackage{color}

\title{\LARGE \bf
Hinted Networks
}

\author{Joel Lamy-Poirier$^{1}$ and Anqi Xu$^{1}$% <-this % stops a space
\thanks{$^{1}$ Element AI, Montreal, Canada
        {\tt\small \{joel,ax\}@elementai.com}}%
}

\begin{document}

\maketitle
\thispagestyle{empty}
\pagestyle{empty}

%-------------------------------------------------------------------------
\begin{abstract}

We present Hinted Networks: a collection of architectural transformations for improving the accuracies of neural network models for regression tasks, through the injection of a prior for the output prediction (i.e. a \emph{hint}). We ground our investigations within the camera relocalization domain, and propose two variants, namely the Hinted Embedding and Hinted Residual networks, both applied to the PoseNet base model for regressing camera pose from an image. Our evaluations show practical improvements in localization accuracy for standard outdoor and indoor localization datasets, without using additional information. We further assess the range of accuracy gains within an aerial-view localization setup, simulated across vast areas at different times of the year.

\end{abstract}

%-------------------------------------------------------------------------
\section{Introduction}

This work targets supervised regression problems, where the aim is to use one source of data as input to correlate and predict a different kind of information as output. The basic statistical approach to regression, i.e. maximum-likelihood estimation, assumes sufficient smoothness of the input-output mapping, meaning that nearby inputs lead to nearby outputs. In practical scenarios however, this assumption is often violated, as the mapping to be learned may vary wildly within a small input region, or even be multi-valued\footnote{These types of non-desired function mappings commonly arise in practice. Notably, learning the inverse of a smooth mapping can be challenging as it may not be injective in general. The camera relocalization problem considered in the present work belongs to this class, as it corresponds to the inverse of the smooth pose-to-image mapping.}. In such scenarios, statistical regression fails at accurately predicting a single output value, and instead returns an average over all possible output hypotheses seen during training. While this \enquote{mode averaging} trait is acceptable in some cases, in other applications a different \enquote{mode seeking} behavior is preferred, which pinpoints one (or ideally all) of the modes of the output hypotheses.

Consider for instance the \enquote{camera relocalization} task, for which the goal is to determine the position and orientation of a camera device given an acquired camera frame. This task can be formulated as supervised regression, however there are no general guarantees of smoothness or bijectivity concerning the image-pose mapping. In fact, an environment might well contain visually-similar landmarks, yet their images with similar appearances could map to drastically different poses. In this case a mode-averaged prediction would localize to somewhere between the similar landmarks, which generally would not be an useful piece of information.

\begin{figure}[t]
\centering
\subfigure[Maximum Likelihood]{
\includegraphics[width=.43\linewidth]{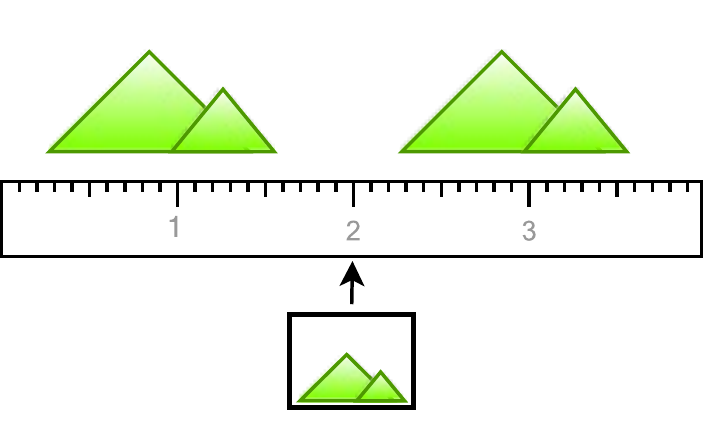}
\label{fig:mode_avg}
}
\hspace{0.02\textwidth}
\subfigure[Hinted Regression]{
\includegraphics[width=.43\linewidth]{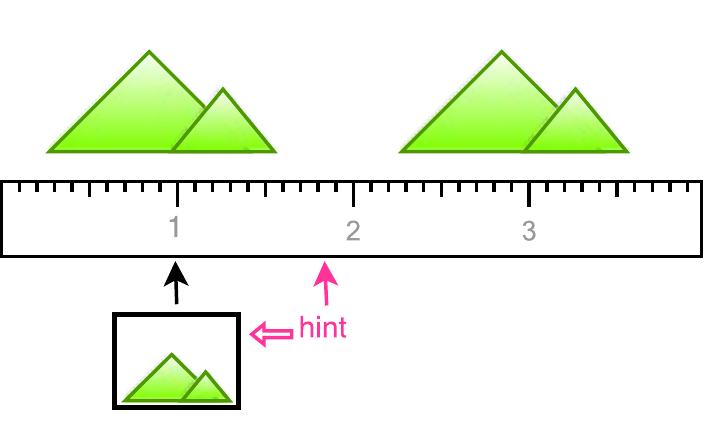}
\label{fig:hinted_mode_seek}
}
\caption{Camera relocalization for a multi-modal scenario: given an ambiguous image, \subref{fig:mode_avg} maximum-likelihood regression will return the average over multiple valid pose hypotheses (i.e. mode averaging), while \subref{fig:hinted_mode_seek} conditioned by even a coarse and noisy hint, our proposed hinted regression models can seek onto a single hypothesis (i.e. mode seeking).}
\label{fig:two_hills}
\end{figure}

As an extreme example, consider a camera observing a 1-D world with two identical hills. Given an image of a single hill, there is insufficient information to disambiguate which of the two landmarks the camera is facing, and the task itself is ill-defined. Moreover, a maximum-likelihood model trained only on single-hill images of both landmarks would predict the pose to be in between the two hills, thus exhibiting \enquote{mode averaging} behavior, as illustrated in Figure~\ref{fig:hinted_mode_seek}.

However, we can still extract useful information from the image by reformulating the problem. For example, we can aim to predict a set of possible positions. Alternatively, given a sufficiently strong prior of the camera's position, we may be able to resolve the ambiguity. Or, even with \textit{any} prior information, we can at least predict its most likely pose. This work aims to realize the last approach.

For the two-hills toy instance, if we have a rough estimate of the camera's position, then we can naturally assume it to be more likely to face the hill closest to this estimate. By changing the regression goal to predict \textit{the most likely position given the estimate}, the image-to-pose mapping will then be deterministically well-defined, with a sole exception when the hint is at the exact center of the two hills. Importantly, even if the estimate is completely random, this model will still produce a correct answer half the time, which might be qualitatively favorable compared to statistical regression being always \enquote{half-wrong}.

More generally, the Hinted Networks transformation reformulates a statistical regression task into a different learning problem with the same inputs and outputs, albeit with an added \enquote{hint} at the input that provides an estimate for the output. While this reformulation works best when a prior exists, our key finding is that Hinted Networks are useful even in the absence of informed priors or auxiliary information. In this work, we present a training procedure that uses synthetically-generated hints at training time, and
\textit{uninformed hints} at query time to improve regression performance.

We devise two strategies for using the hint: \enquote{Hinted Embedding}, which uses the hint as a conditioning prior to resolve ambiguities in the input data, and \enquote{Hinted Residual}, which builds upon Hinted Embedding by converting the learning task from absolute-value into residual regression. A key motivation and benefit of Hinted Networks is their \enquote{mode seeking} behaviors. Also, since the hint and output have the same representation, this feed-forward model can be improved through iterative refinement, by recurrently feeding predictions as successive hints to refine accuracy.

Our investigations of Hinted Networks are grounded in camera relocalization tasks. Such ability to localize camera images is invaluable for diverse applications, for example replacing Global Positioning System (GPS) and Inertial Navigation System (INS), providing spatial reference for Augmented Reality experiences, and helping robots and vehicles self-localize. In these setups, often there are auxiliary data sources that can help with localization, such as GPS and other sensors, as well as temporal odometry priors. Hinted Networks are motivated by the use of auxiliary sources as \textit{informative hints} to facilitate general regression problems.

Hinted Networks build upon several well-established concepts in machine learning. For instance, Hinted Embedding networks apply the hint as a conditioning variable, which is analogous to the use of prior in Bayesian inference~\cite{Koller09PGM}. Nevertheless, our formulation can be seen as a \enquote{poor-man's} alternative to Maximum A Posteriori or full Bayesian inference, as we sacrifice the advantages of learning a probabilistic distribution in favor of a simpler deterministic mapping. Additionally, residual connections such as the ones used in Hinted Residual networks have been shown to often improve training speed and accuracy~\cite{He16ResNet}. Finally, our use of recurrent inference is structurally similar to plain Recurrent Neural Networks (RNN)~\cite{Goodfellow16DL}, although it fulfills a different purpose of iterative refinement.

Our main contribution is to demonstrate improved prediction accuracies for terrestrial localization tasks using Hinted Networks. In particular, we use \textit{uninformed hints} at query time to fairly compare against an appearance-only baseline method, PoseNet~\cite{Kendall15PoseNet,Kendall17PoseNetLearnedLoss}. We also explore the benefits of Hinted Networks for aerial-view localization tasks within simulated large-scale environments, with emphasis on repeated patterns, seasonal variations and high-density training samples. Notably, we find Hinted Networks to significantly outperform PoseNet for this task.

%-------------------------------------------------------------------------
\section{Related Work}

Hinted Networks seek to improve regression performance when using uninformed hints during inference, which are sampled from random noise. Depeweg \textit{et al.}~\cite{Depeweg17NoisyRL} also employed a stochastic conditioning input within a Bayesian Neural Network to enhance dynamics modeling and avoid mode averaging for model-based reinforcement learning tasks.
Separately, Hinted Networks refine hints (by combining with input data) into less noisy target predictions, which is similar to Denoising Autoencoders~\cite{Vincent08DAE}.
Finally, the combination of residual and recurrent connections has also been explored for improving the accuracy of sequence classification tasks~\cite{Wang16RRN,Wang17RRA}.

\subsection{Camera Relocalization}

The ability to predict the position and orientation of a camera image has many applications in Augmented Reality and mobile robotics, and especially complements Visual Odometry (VO) and Simultaneous Localization and Mapping (SLAM) systems by (re-)initializing pose when tracking fails. A survey of solutions for camera relocalization, which is synonymous to visual-based localization, image localization, etc., is provided in~\cite{Piasco18VBLSurvey}. Similar methods have also been applied to the related task of visual place recognition~\cite{Lowry16VPRSurvey}, in which metric localization can be used as a means towards semantic landmark association.

An effective family of localization methods~\cite{Svarm14P3PRansac,Sattler17ActiveSearch} work by matching appearance-based image content with geometric structures from a 3-D environment model. Such models are typically built off-line using Structure from Motion (SfM) and visual SLAM tools (e.g.~\cite{Bundler,VisualSFM,Mur17ORBSLAM2}). Given a query image, point features (e.g. SIFT~\cite{Lowe99SIFT}) corresponding to salient visual landmarks are extracted and matched with the 3-D model, in order to triangulate the resulting pose. While these methods can produce extremely accurate pose estimates, building and maintaining their 3-D environment models is very costly in resources, and this structure-based approach also tends to not generalize well at scale and under appearance changes.

In contrast, PoseNet~\cite{Kendall15PoseNet} is an appearance-only approach for camera relocalization, constituting of a Convolutional Neural Network (CNN) \enquote{backbone}, and separate regressors for predicting position and orientation. While the original PoseNet's training loss used a hand-tuned $\beta$ to balance the different scales for position and orientation, a follow-up work~\cite{Kendall17PoseNetLearnedLoss} replaced $\beta$ with weights that learned the homoscedastic uncertainties for both sources of errors.

Recent work has aimed to address fundamental limitations of PoseNet's appearance-only approach by adding temporal and geometric knowledge. MapNet~\cite{Brahmbhatt18MapNet} incorporated geometry into the training loss by learning a Siamese PoseNet pair for localizing consecutive frames. The MapNet+ extension added relative pose estimates from unlabeled video sequences and other sensors into the training loss, while MapNet+PGO further enforced global pose consistency over a sliding window of recent query frames. Laskar~\textit{et al.}~\cite{Laskar17RelPosePipeline} also learned a Siamese CNN backbone to regress relative pose between image pairs, but then used a memory-based approach for triangulating the query pose from visually similar training samples. Finally, VLocNet~\cite{Valada18VLocNet} added an auxiliary odometry network onto PoseNet to jointly regress per-frame absolute pose and consecutive-pair relative pose, while the follow-up VLocNet++~\cite{Radwan18VLocNetPP} further learns semantic segmentation as an auxiliary task. By sharing CNN backbone weights at early layers, joint learning of absolute pose, relative pose, and auxiliary tasks led to significantly improved predictions, and even surpassing the performance of 3-D structure-based localization in some cases.

\begin{figure*}[t]
\centering
\subfigure[Feed-forward base model]{
\includegraphics[width=.31\linewidth]{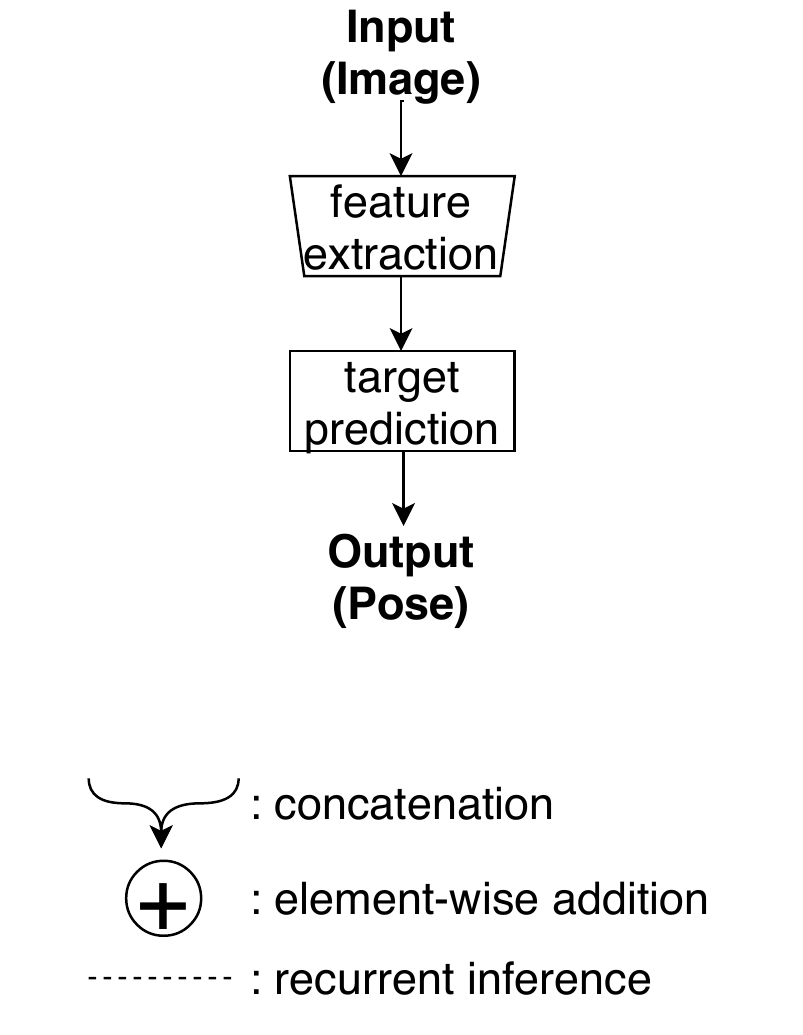}
\label{fig:ff_model}
}
\subfigure[Hinted Embedding model]{
\includegraphics[width=.31\linewidth]{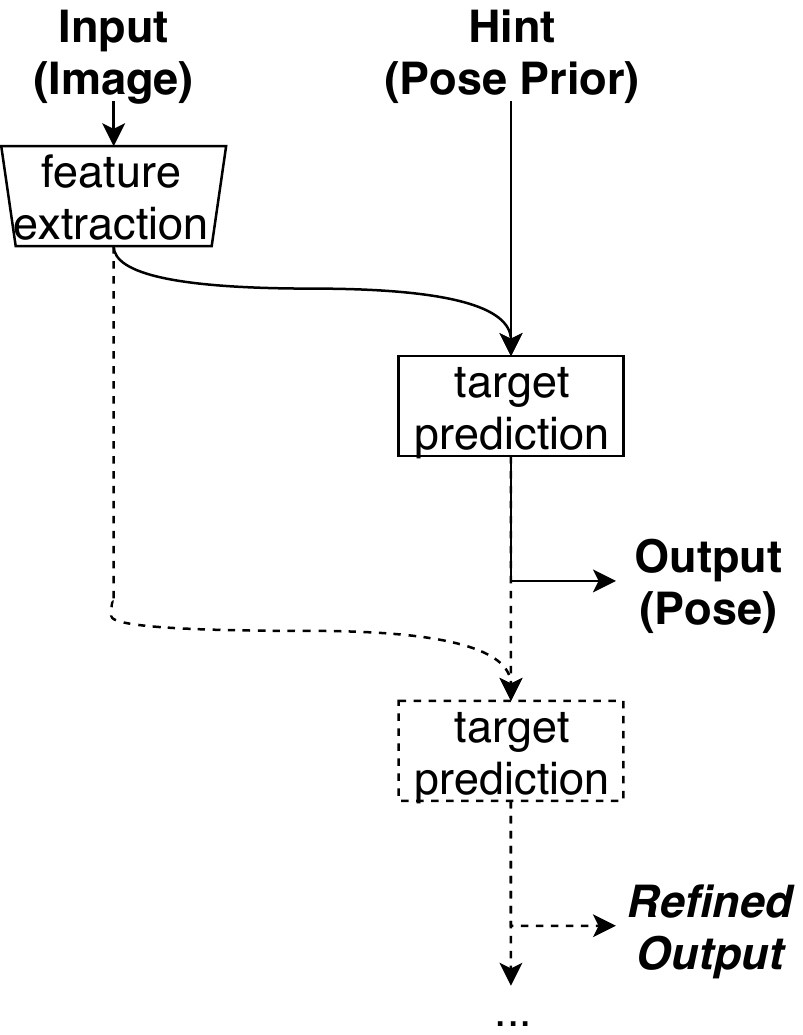}
\label{fig:he_model}
}
\subfigure[Hinted Residual model]{
\includegraphics[width=.31\linewidth]{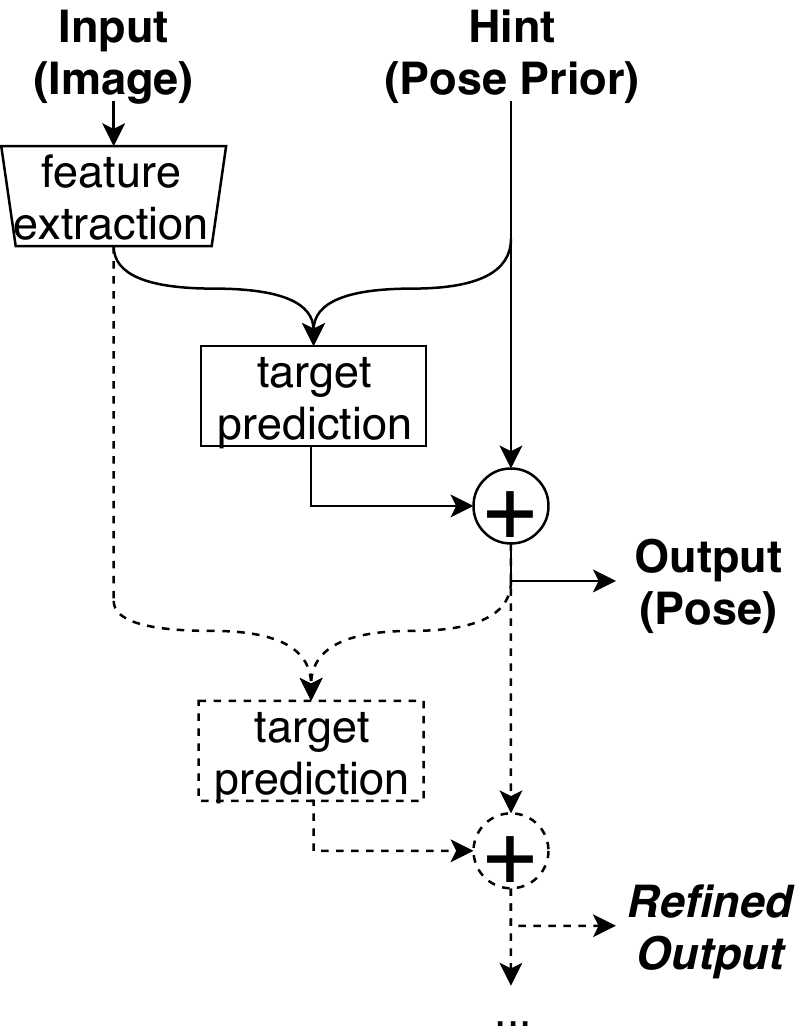}
\label{fig:hr_model}
}
\caption{Layout of network architectures for regression (e.g. camera relocalization) tasks.}
\label{fig:architectures}
\end{figure*}

In contrast to these localization-specific model advances, Hinted Networks represent a weaker yet task-independent way of enhancing appearance-only regression using temporal/geometric information via hints. We also stress that while the above methods require additional information to make predictions, Hinted Networks do not.

Other advances have facilitated model training through data augmentation. SPP-Net~\cite{Purkait17SPPNet} combined a 3-D structure approach with PoseNet by processing a sampled grid of SIFT features through a CNN backbone and pose prediction layers. While this architecture resulted only in comparable performance to PoseNet, notable gains were achieved by training on extra poses, which were synthesized from a 3-D SIFT point-cloud model. More directly, Jia~\textit{et al.}~\cite{Jia16RenderAugmentation} showed improved accuracy by importing a dense SfM environment model into a graphics engine and then training PoseNet on densely-sampled synthesized scenes. We will assess the effects of dense sampling in its limit within an aerial-view localization setup by using an online synthetic data generator.

%-------------------------------------------------------------------------
\section{Hinted Networks}

Hinted Networks are a set of transformations for neural network regression models aimed at simplifying the target prediction task by providing hints of the output values. While some domains offer natural sources of auxiliary information that can be used as \textit{informed hints}, such as using GPS for camera relocalization tasks, we will importantly demonstrate that Hinted Networks can improve prediction accuracy over base models without relying on extra data sources at inference time, through the use of \textit{uninformed hints}.

\subsection{Hinted Architectures}

Many neural networks can be seen as having two parts, as shown in Figure~\ref{fig:ff_model}: one sub-model compresses input data into a set of extracted features (a.k.a. an \enquote{embedding}), and then a second sub-model processes features to predict the target value.
This factorized structure allows the transfer and reuse of pre-trained feature extractors between tasks~\cite{Oquab14Transfer}, which reduces convergence time and allows training with small datasets (e.g.~\cite{Kendall15PoseNet}).

Given a regression task, assume for now that a coarse estimate of the ground-truth target is available, then it can be used as supplementary knowledge to help with prediction. Following the embedding/prediction factorized view above, this prior knowledge can be injected by concatenating the estimate, or \enquote{hint}, with the feature embedding. This \enquote{Hinted Embedding} model is shown in Figure~\ref{fig:he_model}.
Note that the target prediction sub-network requires at least two non-linear layers for the hint and embedding to be mixed sufficiently.

In addition to conditioning the input embedding, we can also add the hint to the output of the target prediction sub-model, as shown in Figure~\ref{fig:hr_model}. This \enquote{Hinted Residual} model transforms the problem from absolute-value regression into residual regression. For image classification tasks, such residual connections have been shown to optimize faster and yield superior accuracy~\cite{He16ResNet}. Especially for camera relocalization, the Hinted Residual model no longer needs to learn to predict \textit{absolute pose} based on an image, but merely \textit{a direction and distance} to move towards the target pose.

\subsection{Training and Inference}

When training a Hinted Network, we sample \textit{informed hints} by applying Gaussian noise around the ground truth value of each data sample. During inference however, we assume no auxiliary information is available, and thus sample \textit{uninformed hints} from an uniform distribution within (estimated) bounds of the environment. While this may appear counterproductive for simple setups like the two-hills scenario in Figure~\ref{fig:two_hills}, our experiments demonstrate that uninformed hints, despite providing much coarser pose estimates than seen during training, help to improve localization accuracy for real-world datasets compared to the PoseNet base model.

Additionally, since the hint and output share the same representation, we can actually feed the predictions back to the network as \textit{subsequent} hints recurrently. Such iterative refinement simplifies the regression task by allowing the model to successively improve prediction accuracies, even when the initial hint is very far from the target. Also, this process is computationally inexpensive, since the feature embedding, whose computation tends to make up the bulk of the workload, can be reused between iterations.

Contrary to RNNs that rely on recurrent training, in this work we apply recurrent hint connections only at inference time, after the model has been trained. A benefit of non-recurrent training is to reduce overfitting by preventing potentially harmful interactions between successive iterations. Additionally, the non-recurrently-trained network observes evenly-spread distributions of priors, in contrast to series of correlated priors observed with an unrolled compute graph.

As a practical consideration, the scale of the noise should be tuned with a bit of care in order to attain optimal inference-time localization accuracy. As extreme cases of failures, if the training hints are too close to the ground truth, the network may choose to output the hint directly and bypass the image-to-pose regression path. On the other hand, if hints are too far away, then the network will not be able to use them efficiently to help disambiguate challenging image-to-pose mapping instances. 

\subsection{Exploration of Alternative Formulations}

As a consequence of our choice of training hint distribution, uninformed hints during inference will likely provide hints much further away from the target than seen during training. Nevertheless, we observed that Hinted Networks were able to generalize. This is for the most part due to recurrent connections, as the networks only need to output a rough pose estimate in the first few iterations. Also, while we experimented with using a heavier-tailed Student's $t$-distribution for sampling uninformed hints at inference time, it did not seem to affect the model's performance.

To justify our choice of non-recurrent training, we investigated also training $N$-step unrolled networks. These exploratory results showed longer training times and larger memory footprints, with \textit{inferior} prediction performances. We attribute this to the fact that the regression tasks across different iterations share an identical nature: notwithstanding the accuracy of the pose hint provided, the model receives the same input frame and must predict the same output pose using the same loss function. In other words, even though there can be some benefits to back-propagating between iterations, for the forward pass the only information transmitted between iterations is the pose prediction itself. Thus at inference time the network is fundamentally symmetric between iterations. Unless that symmetry is removed, e.g. with the transmission of additional variables together with the pose, there can only be minimal benefit to recurrent training.

We further experimented with an interesting use of Hinted Networks: passing through multiple \textit{independent} uniformed hints to predict a \textit{distribution} of output predictions. While this is functionally similar to Bayesian Neural Networks~\cite{Gal16Thesis,Kendall16BayesianPoseNet}, we found that the resulting distribution after multiple iterations collapsed to a small, discrete set of points. This is an expected outcome of Hinted Networks' mode-seeking attribute. We plan to study distributional inference further in the future, possibly by combining with dropout and other Bayesian techniques.

%-------------------------------------------------------------------------
\section{Hinted Networks for Camera Relocalization}

\subsection{PoseNet Baseline Model}

The base model for solving camera relocalization tasks in our grounded investigations of Hinted Networks is derived from PoseNet, and specifically its \enquote{PoseNet2} variant with learned $\sigma^2$ weights for homoscedastic uncertainties~\cite{Kendall17PoseNetLearnedLoss}. This model's architecture is derived from the GoogLeNet (Inception v1) classifier~\cite{Szegedy15InceptionV1}, which is truncated after the final pooling layer. In place of the removed softmax, PoseNet2 attaches a pose prediction sub-network, which is composed of a single $2048$-channel fully-connected hidden layer (with ReLU activation) acting on the feature space, followed by linear output layers for predicting 3-D Cartesian position $\pmb{x}$ and 4-D quaternion orientation $\pmb{q}$.

The per-image ($I$) contribution to PoseNet2's training loss is given as:

\begin{align}
\mathcal{L} \left( I \right) = & \left\Vert \pmb{x} - \pmb{\hat{x}} \right\Vert_\gamma \cdot exp(-\hat{s}_x) + \hat{s}_x + \nonumber\\
& \left\Vert \pmb{q} - \pmb{\hat{q}} \right\Vert_\gamma \cdot exp(-\hat{s}_q) + \hat{s}_q
\end{align}
\noindent where $\pmb{\hat{x}}$ and $\pmb{\hat{q}}$ are the target position and (normalized quaternion) orientation, $\hat{s}_x$ and $\hat{s}_q$ are associated weights for learning each component's data-driven homoscedastic uncertainty (akin to Lagrange multipliers), and $\gamma=2$ is the degree of the \textit{p}-norm.

We re-implemented PoseNet2 using the TensorFlow-Slim library~\cite{TFSlim}, and in particular reused an existing GoogLeNet CNN backbone with pre-trained weights on the \textit{ImageNet} dataset. This model maps $224 \times 224$ color images into a 1024-dimensional feature space. The TF-Slim implementation deviates from the original formulation~\cite{Kendall15PoseNet} by adding batch normalization after every convolutional layer~\cite{Szegedy15InceptionBN}. For simplicity, we omit the auxiliary branches from the Inception v1 backbone.

\subsection{Hinted PoseNets Implementation Details}

As seen in Figure~\ref{fig:architectures}, both the Hinted Embedding and Hinted Residual models extend from the base model's architecture. While the feature extractor is copied verbatim from PoseNet2's CNN backbone, the target prediction sub-network for hinted models needs to be modified in order to accept the concatenation of the feature vector and the hint vector as input. Thus, the hinted sub-network contains \textit{three} hidden layers (as opposed to PoseNet2's single $2048$-channel pose prediction layer), with $1024$, $2048$, and $1024$ channels respectively, to ensure sufficient mixing of the concatenated tensor. While the increased network capacity may seem to give an unfair advantage to hinted architectures, we tested the proposed three-layer pose predictor on the PoseNet base model and found that it performed worse than its original configuration. This is consistent with previous reports that PoseNet's pose prediction sub-network is already prone to overfitting~\cite{walch17PoseNetLSTM}.

We pre-process images by down-scaling and square-cropping to a resolution of $224 \times 224$, and then normalizing pixel intensities to range from $-1$ to $1$.
We also normalize and sign-disambiguate all target quaternions by restricting them to a single hyper-hemisphere.

Since each orientation can be represented ambiguously by two sign-differing quaternion vectors, it is crucial for both training and evaluating regressors to consistently map all quaternions onto a single hyper-hemisphere. This is achieved by unit-normalizing their magnitudes, and also sign-normalizing the first non-zero component.

Furthermore, contrary to other PoseNet-style systems (e.g.~\cite{Kendall17PoseNetLearnedLoss,Brahmbhatt18MapNet,Valada18VLocNet}), we train models on PCA-whitened~\cite{Duda12PR} representations of both position and orientation.
In addition to normalizing across mismatched dimensions, whitening removes the need to manually specify initial scales for regression-layer weights and for hints. Having initial pose estimates matching the scales of each environment is crucial during training~\cite{Kendall15PoseNet}. Predicted poses are de-whitened prior to evaluating the training loss and at query time.

%-------------------------------------------------------------------------
\section{Terrestrial Localization Experiments}
\label{sec:terr_loc}

We now evaluate Hinted Networks for camera relocalization tasks on the outdoor \textit{Cambridge Landmarks}~\cite{Kendall15PoseNet} dataset and indoor \textit{7-Scenes}~\cite{Shotton137Scenes} dataset. These \textit{terrestrial} datasets are comprised of images that were taken using hand-held cameras, targeting nearby landmarks with predominantly forward-facing orientations. By contrast, the next section will present evaluations within the separate domain of \textit{aerial-view} localization, where the goal will be to localize high-altitude downward-facing camera frames acquired by aerial drones.

All models are optimized with Adam~\cite{Kingma15Adam} using default parameters and a learning rate of $1 \times 10^{-4}$, for $50$k (\textit{7-Scenes}) and $100$k (\textit{Cambridge}) iterations, with a batch size of $64$. During training, hints are sampled from Gaussian noise around ground truth with uncorrelated deviations of $0.3$ along each PCA-whitened axis. During inference, hints are initialized with a unit-scale normal distribution and fed through the network until convergence.

\subsection{Comparison of Localization Performance}

As shown in Table~\ref{tab:terr_loc_results}, our PoseNet2 implementation attains slightly worse localization accuracy compared to~\cite{Kendall17PoseNetLearnedLoss}. We attribute this difference to minor discrepancies in the architecture, pre-processing, and training regime. To isolate the effects of the architecture from those of experimental setup, we use our implementation as the comparative baseline.

The Hinted Residual network boasts superior localization accuracies for most scenes, both in position and especially in orientation. On the other hand, the Hinted Embedding network converges to pose predictions that are no better than PoseNet2 for most scenes. Furthermore, the test-set errors for certain scenes are slightly elevated during late training for PoseNet2, thus reflecting a sign of overfitting; this is not observed for neither hinted architectures.

\begin{table*}[htb]
\centering
\begin{tabular}{llll|l|lll}
\textit{Cambridge} & Training & Test& Size & PoseNet2 & PoseNet2 & Hinted & Hinted \\
\textit{Landmarks}~\cite{Kendall15PoseNet}& Samples& Samples& & (from \cite{Kendall17PoseNetLearnedLoss}) & (our baseline) & Embedding & Residual \\
\hline
\hline
King's College & 1220 & 343 & 5600$m^2$ & $0.99$m, $1.06^\circ$ & $1.30$m, $1.78^\circ$ & $1.67$m, $2.22^\circ$ & $\mathbf{0.95}$m, $\mathbf{1.32}^\circ$ \\
Old Hospital & 895 & 182 & 2000$m^2$ & $2.17$m, $2.94^\circ$ & $2.29$m, $3.72^\circ$ & $2.41$m, $4.95^\circ$ & $\mathbf{2.02}$m, $\mathbf{3.03}^\circ$ \\
Shop Facade & 231 & 103 & 875$m^2$ & $1.05$m, $3.97^\circ$ & $1.47$m, $5.45^\circ$ & $\mathbf{1.36}$m, $5.67^\circ$ & $1.53$m, $\mathbf{5.13}^\circ$ \\
St Mary's Church & 1487 & 530 & 4800$m^2$ & $1.49$m, $3.43^\circ$ & $1.95$m, $5.49^\circ$ & $2.09$m, $5.85^\circ$ & $\mathbf{1.83}$m, $\mathbf{4.93}^\circ$ \\
Great Court & 1532 & 760 & 8000$m^2$ & $7.00$m, $3.65^\circ$ & $6.78$m, $5.06^\circ$ & $\mathbf{4.98}$m, $4.84^\circ$ & $6.75$m, $\mathbf{4.49}^\circ$ \\
\hline 
Average &  &  &  & $2.54$m, $3.01^\circ$ & $2.76$m, $4.30^\circ$ & $\mathbf{2.50}$m, $4.70^\circ$ & $2.62$m, $\mathbf{3.78}^\circ$ \\
 & & & & \\
\textit{7-Scenes}~\cite{Shotton137Scenes} & & & & \\
\hline
\hline
Chess & 4000 & 2000 & 6.0$m^3$ & $0.14$m, $4.50^\circ$ & $0.134$m, $\phantom{0}5.96^\circ$ & $0.153$m, $\phantom{0}7.13^\circ$ & $\mathbf{0.115}$m, $\mathbf{\phantom{0}5.04}^\circ$ \\
Fire & 2000 & 2000 & 2.5$m^3$ & $0.27$m, $11.8^\circ$ & $0.281$m, $11.42^\circ$ & $0.292$m, $12.72^\circ$ & $\mathbf{0.279}$m, $\mathbf{10.91}^\circ$ \\
Heads & 1000 & 1000 & 1.0$m^3$ & $0.18$m, $12.1^\circ$ & $0.153$m, $\mathbf{13.22}^\circ$ & $\mathbf{0.148}$m, $13.55^\circ$ & $0.167$m, $13.51^\circ$ \\
Office & 6000 & 4000 & 7.5$m^3$ & $0.20$m, $5.77^\circ$ & $0.211$m, $\phantom{0}7.42^\circ$ & $0.237$m, $\phantom{0}8.11^\circ$ & $\mathbf{0.196}$m, $\mathbf{\phantom{0}6.69}^\circ$ \\
Pumpkin & 4000 & 2000 & 5.0$m^3$ & $0.25$m, $4.82^\circ$ & $0.258$m, $\phantom{0}6.12^\circ$ & $0.266$m, $\phantom{0}7.34^\circ$ & $\mathbf{0.229}$m, $\mathbf{\phantom{0}5.28}^\circ$ \\
Red Kitchen & 7000 & 5000 & 18.0$m^3$ & $0.24$m, $5.52^\circ$ & $0.273$m, $\phantom{0}6.84^\circ$ & $0.276$m, $\phantom{0}7.68^\circ$ & $\mathbf{0.252}$m, $\mathbf{\phantom{0}6.34}^\circ$ \\
Stairs & 2000 & 1000 & 7.5$m^3$ & $0.37$m, $10.6^\circ$ & $0.300$m, $12.90^\circ$ & $0.355$m, $13.09^\circ$ & $\mathbf{0.271}$m, $\mathbf{11.79}^\circ$ \\
\hline 
Average &  &  &  & $0.24$m, $7.87^\circ$ & $0.230$m, $\phantom{0}9.13^\circ$ & $0.247$m, $\phantom{0}9.95^\circ$ & $\mathbf{0.215}$m, $\mathbf{\phantom{0}8.51}^\circ$ \\
\end{tabular}
\caption{Median localization results for terrestrial datasets.}
\label{tab:terr_loc_results}
\end{table*}

\begin{figure*}[ht]
\centering
\includegraphics[width=.95\linewidth]{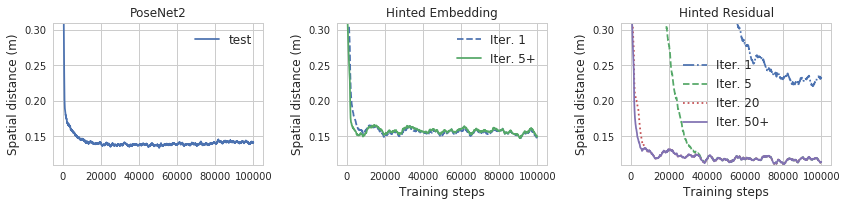}
\caption{Convergence of median test-set spatial errors for the Chess scene.}
\label{fig:convergence_plot}
\end{figure*}

Additionally, as seen in Figure~\ref{fig:convergence_plot}, while PoseNet2 models train faster, Hinted Residual networks are able to leverage iterative refinement to make more accurate predictions. In contrast, after an initial learning phase, Hinted Embedding models converge numerically without benefiting from recurrent connections.

These findings suggest that the use of an uninformed hint \textit{solely as a prior} does not help with pose prediction and instead complicates learning with added parameters. On the other hand, the addition of recurrent connections significantly improves prediction accuracies. We thus conclude that the main advantage of hinted architectures, at least using uninformed hints, is not the hints themselves but rather the iterative refinement process.

\subsection{Effects of Noise Scale for Training-Time Hints}

For these experiments, we have arbitrarily set the Gaussian noise scale for hints during training to be $0.3 \times$ of each PCA-normalized axis. Still, Hinted Networks are mostly insensitive to this hyper-parameter, as long as it lies within a reasonable range, as seen in Figure~\ref{fig:convergence_plot}. Naturally, when the hint noise is near zero, hinted models achieve near-perfect \textit{training} performance by simply passing the hint as output, which leads to poor \textit{test-time} generalization. Perhaps surprisingly however, the Hinted Residual model still outperforms the baseline even when the training hint is extremely noisy, thus carrying little information. This finding again highlights the importance of recurrence inference in our hinted architectures. In contrast, Hinted Embedding models cannot make good use of the hint, and even perform better when trained with uninformative hints.

\begin{figure*}[htb]
\centering
\includegraphics[width=.76\linewidth]{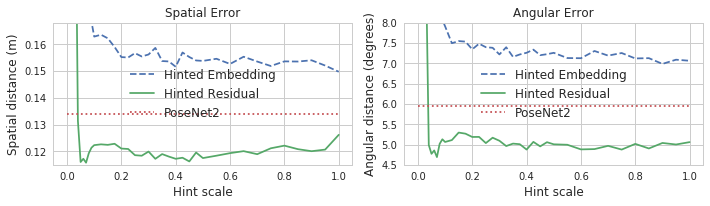}
\caption{Effect of hint scale on test-set errors for the Chess scene after $25$k iterations.}
\label{fig:hint_scale_plot}
\end{figure*}

%------------------------------------------------------------------------
\section{Aerial-View Localization Experiments}
\label{sec:air_loc}

Hinted Networks are designed to facilitate learning of challenging regression instances with significant self-similarities among input values. To explicitly assess the range of potential performance gains, we conduct a set of localization experiments on aerial views, for which we can generally expect more visual ambiguities. Such aerial-view localization is useful in diverse GPS-denied scenarios, including underwater and extra-terrestrial planetary surfaces.

We train and evaluate models on synthesized downward-facing images from aerial drones, which are extracted from large-scale satellite imagery. This setup is motivated both by data availability and the possibility to deliberately factor out effects of sparse sampling and limited dataset size by using online data generators. We acknowledge that this setup does not fully reflect real-world conditions, and thus model deployment would likely benefit from domain adaptation and/or datasets of actual footage from aerial drones.

\begin{table*}[ht]
%\scriptsize
\centering
\begin{tabular}{|c|c|c|c|c|c|c|}
\hline
Region & Tile & Location & Features & Seasons & Training images & Test images \\
\hline
Al-Ahsa & 39QXD & Saudi Arabia & Desert, Sand & None & 5 & 5 \\
Beijing & 50TMK & China & Urban, Mountains & Dry, Wet & 7 & 3 \\
Death Valley & 11SMA & United States & Desert, Hills & None & 5 & 4 \\
Finke Gorge & 52KHU & Australia & Desert, Hills & None & 7 & 4 \\
Montreal & 18TXR & Canada & Urban, Rural, Forest & Summer, Winter & 6 & 4 \\
Moscow & 37UDB & Russia & Urban, Rural, Forest & Summer, Winter & 4 & 2 \\
Tokyo & 54SUE & Japan & Urban, Mountains, Water & Summer, Mild Winter & 9 & 4 \\
\hline
\end{tabular}
\caption{Visually-distinct regions chosen for aerial-view localization experiments.}
\label{tab:sat_env}
\end{table*}

\subsection{Setup}

The satellite scenes used are based on data from the Sentinel-2 Earth observation mission~\cite{Sentinel} by the European Space Agency (ESA). All imagery is publicly and freely available on ESA's Copernicus Open Access Hub~\cite{Copernicus}. We select seven regions with various degrees of self-similarity and seasonal variations, which are enumerated along with their main features in Table~\ref{tab:sat_env}. Each region maps to a specific \textit{Sentinel-2} mission tile, and covers a square area of $12,000$\,km$^2$, with a pixel resolution of $10$\,m. For each region, we choose up to thirteen non-cloudy sample images, depending on availability, and split the dataset into between 4 to 9 training and between 2 to 5 test tiles. Samples were taken at various times of the year between $2016$ and $2018$. A sample image is shown in Figure~\ref{fig:sample_tile}\footnote{Please see supplementary material for a comprehensive list of product IDs, as well as a sample tile image per region.}. While we aim to split datasets randomly, we also ensure that each season is represented in both the training and test sets. We further experiment with variations of these setups to study effects of altitude ranges, cross-seasonal variations, and the presence of clouds, as enumerated in Table~\ref{tab:sat_exp}.

\begin{table*}[ht]
\centering
\begin{tabular}{|c|c|c|c|c|c|}
\hline
Setup & Region & Features & Training images & Test images \\
\hline
Low Altitudes & Montreal & Low altitude range, from $1$\,km to $2$\,km  & 5 & 5 \\
High Altitudes & Montreal & High altitude range, from $3$\,km to $5$\,km  & 5 & 5 \\
Wide Altitudes & Montreal & Wide altitude range, from $1$\,km to $5$\,km  & 5 & 5 \\
Winter & Montreal & Trained and evaluated on winter views only & 2 & 2 \\
Winter$^*$ & Montreal & Trained on all scenes, evaluated on winter views only & 2 & 4 \\
Summer & Montreal & Trained and evaluated on summer views only & 4 & 2 \\
Summer$^*$ & Montreal & Trained on all scenes, evaluated on summer views only & 4 & 4 \\
Clouds & Al-Ahsa & Trained and evaluated on partially cloudy (0-10\%) views & 5 & 2 \\
\hline
\end{tabular}
\caption{Variations on the chosen regions selected for experimentation.}
\label{tab:sat_exp}
\end{table*}

\begin{figure*}[ht]
\centering
\subfigure{
\includegraphics[width=.18\linewidth]{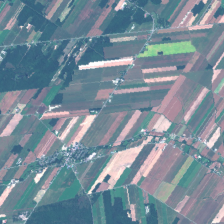}
\label{fig:sample_gen_0}
}
\subfigure{
\includegraphics[width=.18\linewidth]{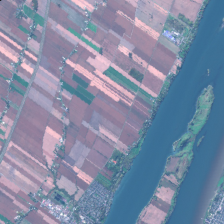}
\label{fig:sample_gen_1}
}
\subfigure{
\includegraphics[width=.18\linewidth]{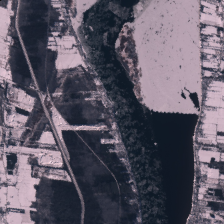}
\label{fig:sample_gen_2}
}
\subfigure{
\includegraphics[width=.18\linewidth]{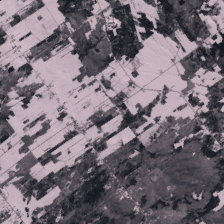}
\label{fig:sample_gen_3}
}
\subfigure{
\includegraphics[width=.18\linewidth]{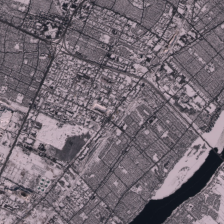}
\label{fig:sample_gen_4}
}
\subfigure{
\includegraphics[width=.18\linewidth]{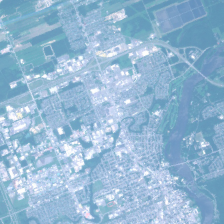}
\label{fig:sample_gen_5}
}
\subfigure{
\includegraphics[width=.18\linewidth]{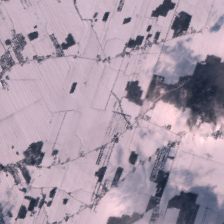}
\label{fig:sample_gen_6}
}
\subfigure{
\includegraphics[width=.18\linewidth]{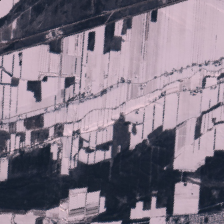}
\label{fig:sample_gen_7}
}
\subfigure{
\includegraphics[width=.18\linewidth]{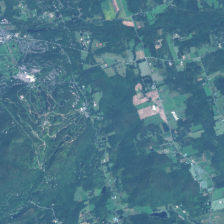}
\label{fig:sample_gen_8}
}
\subfigure{
\includegraphics[width=.18\linewidth]{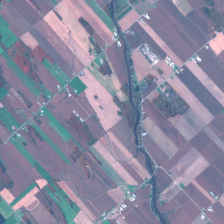}
\label{fig:sample_gen_9}
}
\caption{Training samples from the Montreal dataset.}
\label{fig:sample_gen}
\end{figure*}

Tile images are converted from 16-bits to 8-bits (per channel) according to pixel intensity ranges. Our data generator synthesizes orthogonally-projected camera frames by uniformly sampling at different positions, orientations, and altitudes, with a horizontal field-of-view of $100^\circ$. Unless otherwise specified, altitudes are sampled between $2$\,km and $3$\,km. Samples of generated images are illustrated in Figure~\ref{fig:sample_gen}.

Model architectures and training regimes are nearly identical to those from the previous section, with the following exceptions. Since each pose only has a single planar yaw angle, we regress a $2$-D cosine-sine heading vector instead of a quaternion\footnote{Similar to previously, headings are normalized for ground truth values during training, but not for predictions.}. We also regress altitude separately from lateral coordinates given their large differences in scale, using an independently-learned uncertainty factor $\hat{s}_z$. Moreover, given the unlimited number of image-pose samples that are generated on-the-fly, models can benefit from longer training, which we set at $500$k iterations.

As a final distinction, for these experiments we set the training hint noise scale to $0.2$ for the spatial dimensions and $0.5$ for the angular ones. As justification, we empirically noted that for certain scenes, angular errors tend to remain high for the hinted architectures unless the hint scale is sufficiently large. Our assessments suggest that the orientation regressor fails catastrophically, resulting in two clusters of errors around $0^\circ$ and $180^\circ$. We thus suspect that these models learn the grid-like structure of some environments (e.g. a lattice of farmlands) instead of the actual orientation.

\subsection{Results}

Table~\ref{tab:air_loc_results} shows the localization performances for the diverse setups. Similar to our terrestrial experiments, Hinted Residual networks demonstrate overall greater accuracy in predicting positions as compared to baseline models, although they do not localize orientation as well. On the other hand, Hinted Embedding models interestingly excel at predicting angular poses. This suggests that hints are especially useful to localize scenes with high degrees of visual ambiguity.

\begin{table*}[ht]
\centering
\begin{tabular}{l|lll}
Region / & PoseNet2 & Hinted & Hinted \\
Setup & (baseline) & Embedding & Residual \\
\hline
\hline
Al-Ahsa & $1824$m, $0.90^\circ$ & $1476$m, $\mathbf{0.89}^\circ$ & $\mathbf{1229}$m, $0.95^\circ$ \\
Beijing & $1047$m, $1.15^\circ$ & $1234$m, $\mathbf{0.99}^\circ$ & $\mathbf{536}$m, $1.14^\circ$ \\
Death Valley & $1152$m, $1.70^\circ$ & $1261$m, $\mathbf{1.23}^\circ$ & $\mathbf{469}$m, $1.62^\circ$ \\
Finke Gorge & $1064$m, $1.09^\circ$ & $1158$m, $\mathbf{1.00}^\circ$ & $\mathbf{387}$m, $1.06^\circ$ \\
Montreal & $1274$m, $1.26^\circ$ & $1338$m, $\mathbf{0.90}^\circ$ & $\mathbf{533}$m, $1.14^\circ$ \\
Moscow & $2403$m, $2.41^\circ$ & $1628$m, $\mathbf{1.51}^\circ$ & $\mathbf{766}$m, $2.20^\circ$ \\
Tokyo & $1314$m, $2.17^\circ$ & $1431$m, $\mathbf{1.61}^\circ$ & $\mathbf{562}$m, $1.94^\circ$ \\
 & & & \\
\hline 
Average & $1440$m, $1.53^\circ$ & $1361$m, $\mathbf{1.16}^\circ$ & $\mathbf{640}$m, $1.44^\circ$ \\
\hline 
Low Alt. & $1815$m, $1.36^\circ$ & $1724$m, $\mathbf{1.07}^\circ$ & $\mathbf{835}$m, $1.53^\circ$ \\
High Alt. & $1045$m, $1.03^\circ$ & $1104$m, $\mathbf{0.77}^\circ$ & $\mathbf{449}$m, $1.01^\circ$ \\
Wide Alt. & $1266$m, $1.19^\circ$ & $1339$m, $\mathbf{0.93}^\circ$ & $\mathbf{620}$m, $1.37^\circ$ \\
Winter & $4444$m, $2.91^\circ$ & $2212$m, $\mathbf{1.50}^\circ$ & $\mathbf{1935}$m, $2.94^\circ$ \\
Winter$^*$ & $1613$m, $1.44^\circ$ & $1363$m, $\mathbf{0.98}^\circ$ & $\mathbf{593}$m, $1.26^\circ$ \\
Summer & $1228$m, $1.15^\circ$ & $1226$m, $\mathbf{0.86}^\circ$ & $\mathbf{570}$m, $1.12^\circ$ \\
Summer$^*$ & $1084$m, $1.11^\circ$ & $1245$m, $\mathbf{0.85}^\circ$ & $\mathbf{484}$m, $1.04^\circ$ \\
Clouds & $9282$m, $\mathbf{1.66}^\circ$ & $9296$m, $1.70^\circ$ & $\mathbf{7336}$m, $1.91^\circ$ \\
\end{tabular}
\caption{Median localization results for aerial-view localization experiments.}
\label{tab:air_loc_results}
\end{table*}

Focusing on the altitude experiments, we unsurprisingly find that all models perform better at high altitudes due to wider camera swaths. We also observe hints to be more useful at lower altitudes given pronounced visual ambiguities, and that the network is capable of learning location-pertinent visual attributes within a wide range of altitudes.
As for cross-seasonal experiments, we find that all networks, independently of their architectures, are able to learn seasonal variations, but more importantly can also leverage data from one season to improve predictions within another view. For instance, including Summer scenes in the Montreal dataset drastically improves prediction accuracy in the harder Winter scenes, even though latter landmark textures are blanketed by snow.

Finally, we find that PoseNet-style models do not perform well in the presence of even relatively small amounts of clouds. We believe that networks fail to learn that clouds are irrelevant to the pose, and instead memorize cloud locations from the training set. This failure is heightened by the small number of training scenes with static cloud placements, and can potentially be addressed by training on other synthetic cloudy scenes.

\begin{figure*}[ht]
\centering
\scriptsize
\includegraphics[width=.85\linewidth]{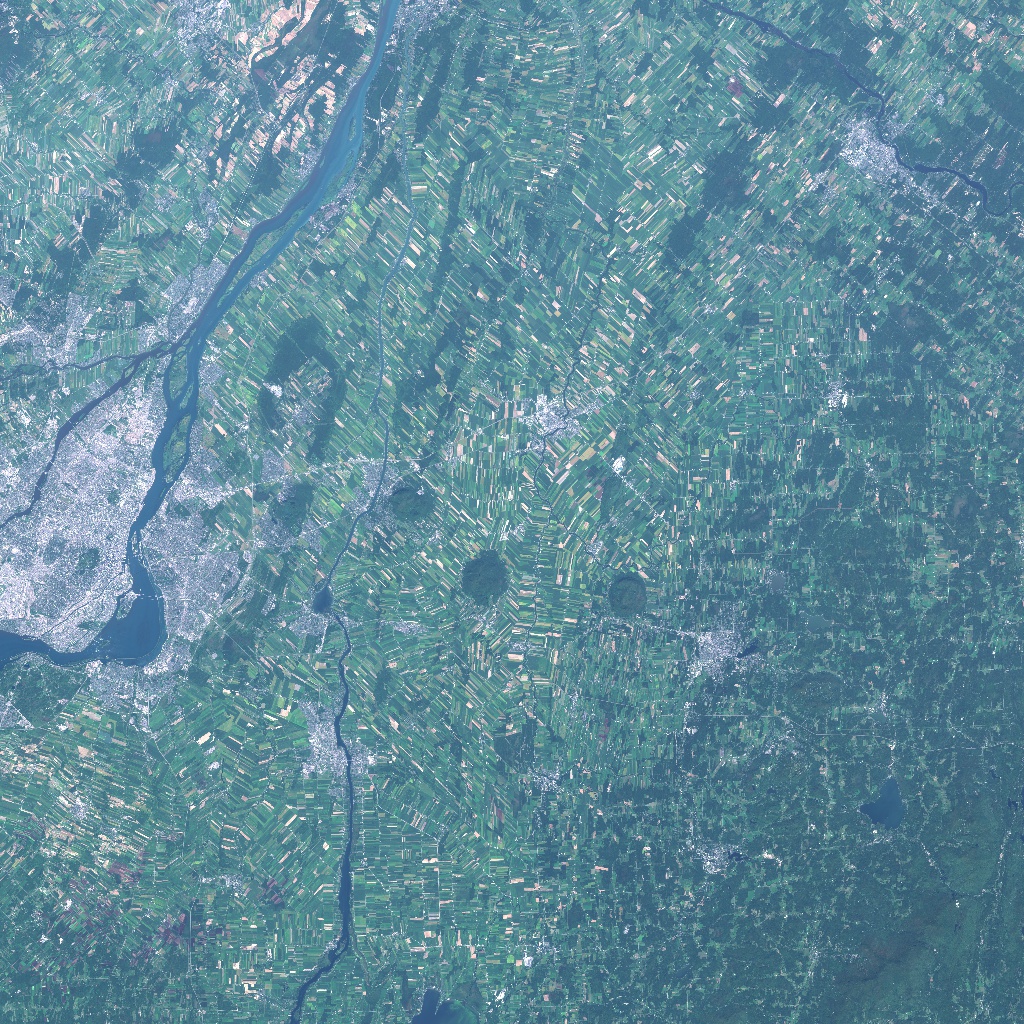}
\caption{A sample tile from the Montreal dataset.}
\label{fig:sample_tile}
\end{figure*}

%------------------------------------------------------------------------
\section{Conclusions}

In summary, Hinted Networks improve the performance of neural network regression models by incorporating a hint of the target as stochastic input. These architectural transformations both simplify the learning task and increase prediction accuracy, through the application of prior conditioning, residual connections, and recurrent connections. The Hinted Residual PoseNet model yields improved performance for camera relocalization tasks compared to the baseline network, for both standard outdoor and indoor terrestrial datasets. We also explored aerial-view localization tasks, for which Hinted Networks showed a widened range of performance gains.

In future work, we want to enhance the sampling of hints during training by adapting and learning distributional scale and shape to match inference-time conditions. We would also like to continue exploring the potential benefits of recurrent hinted models during training, and of informed hints during inference when auxiliary data sources are available. We are further interested in using multiple uninformed hints to predict a target \textit{distribution}, possibly by combining with dropout and other Bayesian techniques~\cite{Gal16Thesis}. Finally, since the hinted network architecture is not specific to camera relocalization tasks, we are excited to assess other potential applications of such regression models.

%------------------------------------------------------------------------
\bibliographystyle{IEEEtran}
\bibliography{hinted_network}

\end{document}